\documentclass{article}
\usepackage{graphicx}
\usepackage{arxiv}
\usepackage{url}
\usepackage{bm}

\title{Addressing Gap between Training Data and Deployed Environment by On-Device Learning}

\begin{document}

\author{
  Kazuki Sunaga\\
  Keio University\\
  3-14-1 Hiyoshi, Kohoku-ku, Yokohama, Japan\\
  \texttt{sunaga@arc.ics.keio.ac.jp}\\
  \And
  Masaaki Kondo\\
  Keio University\\
  3-14-1 Hiyoshi, Kohoku-ku, Yokohama, Japan\\
  \texttt{kondo@acsl.ics.keio.ac.jp} \\
  \And
  Hiroki Matsutani\\
  Keio University\\
  3-14-1 Hiyoshi, Kohoku-ku, Yokohama, Japan\\
  \texttt{matutani@arc.ics.keio.ac.jp}\\
}
\maketitle

\begin{abstract}
The accuracy of tinyML applications is often affected by various
environmental factors, such as noises, location/calibration of
sensors, and time-related changes.
This article introduces a neural network based on-device learning
(ODL) approach to address this issue by retraining in deployed environments.
Our approach relies on semi-supervised sequential training of multiple
neural networks tailored for low-end edge devices.
This article introduces its algorithm and implementation on
wireless sensor nodes consisting of a Raspberry Pi Pico and low-power
wireless module.
Experiments using vibration patterns of rotating machines demonstrate
that retraining by ODL improves anomaly
detection accuracy compared with a prediction-only deep neural network
in a noisy environment.
The results also show that the ODL approach can save
communication cost and energy consumption for battery-powered Internet
of Things devices.
\end{abstract}

%
%
\keywords{Machine learning \and Neural network \and Edge AI \and
       On-device learning \and OS-ELM \and Wireless sensor node}

\maketitle

\section{Challenges on Edge AI}
\begin{figure*}[t]
\centerline{\includegraphics[width=1.0\linewidth]{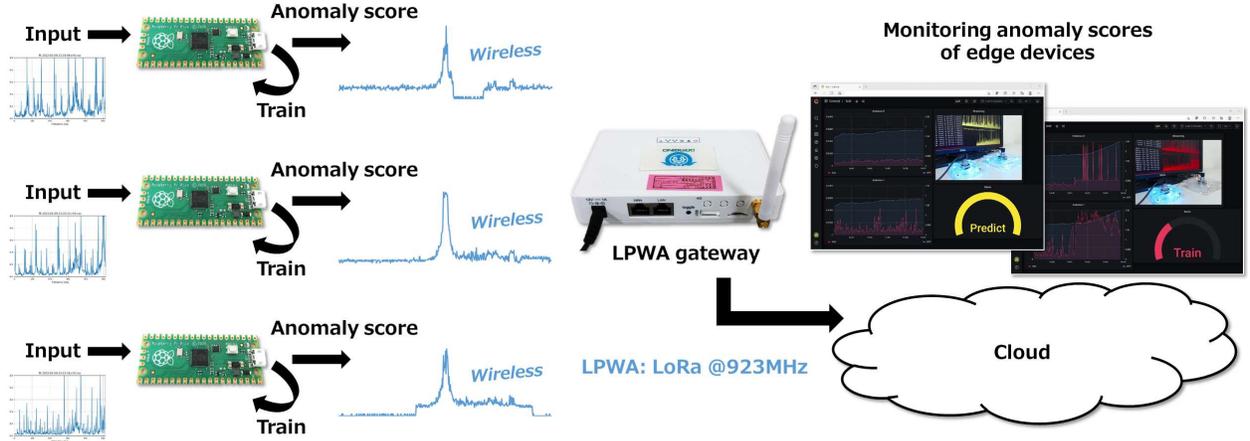}}
\caption{Edge-cloud AI system where sensing, preprocessing,
  prediction, and training are done by edge devices (left) while their
  monitoring and controlling interfaces are provided by the cloud
  (right).
  Edge devices transmit prediction results to a gateway wirelessly.}
\label{fig:edge-cloud}
\end{figure*}

Edge intelligence has gained significant attention
due to the proliferation of emerging Internet of Things (IoT) and
artificial intelligence (AI) technologies \cite{IEEE2019_Zhou}.
Most edge AI applications are generally viewed as an edge-cloud
cooperative system, as shown in Figure \ref{fig:edge-cloud}.
Classical edge applications are usually responsible for
sensing environmental data and sending them to the cloud.
Due to the wide spread of edge AI technologies, AI inference is
becoming a major task of edge devices.

These ``prediction-only edge AI'' systems work properly if the target
environment around the edge devices is fairly static.
However, an environment incessantly changes over time in the real world.
For example, a vibration pattern of some equipment captured by
an IoT sensor would be affected by the setting of a room ventilation
fan, as demonstrated later in this article.
This problem is known as the concept drift \cite{TKDE2019_Lu}.
An on-device learning (ODL) approach \cite{TC2020_Tsukada}, in which edge
devices learn the target phenomenon by themselves, is practical
to address this problem without demanding an excessive
generalization performance.

\begin{figure*}[t]
\centerline{\includegraphics[width=1.0\linewidth]{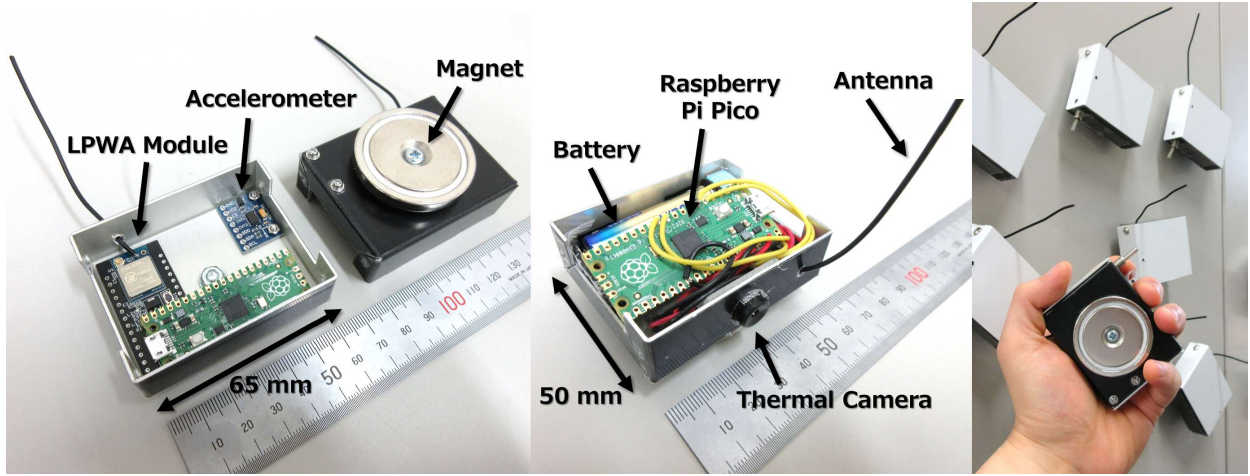}}
\caption{Implementations of wireless sensor nodes equipped with
  accelerometer (left) and thermal camera (center).
  Those with an accelerometer are used in this article.
  They are battery-powered and attached to target objects using a magnet
  (right).}
\label{fig:prototype}
\end{figure*}

Prediction-only edge AI needs a high generalization performance 
to achieve good accuracy wherever it is deployed.
However, this is not necessarily important in the ODL scenario since
it is trained in a deployed environment and operated in the same
environment unless concept drifts occur.
This enables us to use low-cost neural networks as machine learning
models if they can be trained in deployed environments.
In this article, we extend our previous work \cite{TC2020_Tsukada} 
to implement multiple neural networks for complex patterns.
We then demonstrate that they can be trained in relatively low-end IoT
devices, such as a Raspberry Pi Pico (Figure \ref{fig:prototype}), using
an online sequential learning algorithm tailored for such devices.

In the ODL approach, we do not upload raw sensing data
to a cloud server except for prediction results.
This is one of the most important characteristics, especially for
wireless sensor nodes since they are mostly restricted by energy
budgets and wireless communication dominates their energy consumption.
Reducing the energy consumed in wireless communication is one of the
perpetual demands for IoT devices.
In this article, we demonstrate that the ODL approach is
beneficial for reducing the energy consumption (and thereby increasing the
battery life) of wireless sensor nodes that use long range (LoRa) as a
low-power wide-area (LPWA) network technology.

\section{ON-DEVICE LEARNING}

In \cite{IEEE2019_Zhou}, edge AI systems are classified into six
levels from ``cloud-edge coinference and cloud training'' (Level 1) to
``all on-device'' (Level 6).
A typical prediction-only edge AI system corresponds to Level 3
while our ODL approach is Level 6.
As the level increases, the data transmission latency and network
bandwidth decrease, while the data privacy and computation cost at
edge increase.
Thus, reducing the computation cost is a primary concern in the
ODL scenario, and it is addressed by redesigning the
machine learning algorithm as introduced in this section.

\begin{figure*}[t]
\centerline{\includegraphics[width=1.1\linewidth]{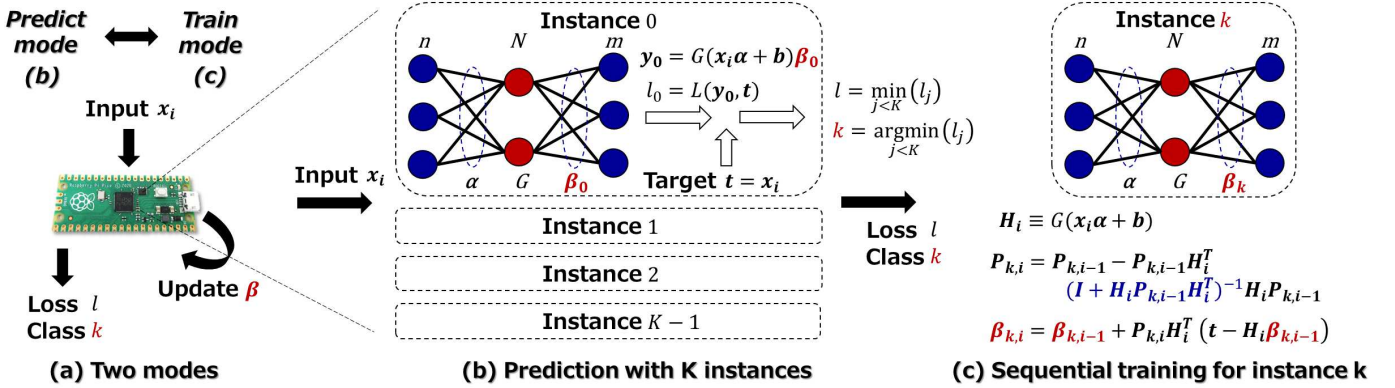}}
\caption{Prediction and sequential training algorithms of ODL
  with multiple instances, where $\bm{x}$ is input data, $l$
  is anomaly score, $k$ is output class, $\bm{\alpha}$ and
  $\bm{\beta}$ are weight parameters, $\bm{b}$ is bias vector,
  $\bm{P}$ is temporary vector, $G$ is activation function, $\bm{H}$
  is hidden layer output, $L$ is loss function, $n$ is number of input
  nodes, $N$ is number of hidden layer nodes, $m$ is number of output
  nodes, and $K$ is number of ODL instances.
  Predict and train modes are switched manually or automatically.}
\label{fig:algorithm}
\end{figure*}

A de facto approach to train artificial neural networks is using a
backpropagation algorithm combined with a stochastic gradient descent
(SGD) or its variant.
A batch of training samples is processed at a time assuming that an
entire training dataset is available.
This assumption is infeasible in low-end IoT devices with limited
memory capacity (e.g., 264 KiB in a Raspberry Pi Pico).
In fact, it is reasonable that sensor data is generated continuously
as a stream.

OS-ELM \cite{TNN2006_Liang} takes a different approach to sequentially
train artificial neural networks.
It is an online sequential learning algorithm for feedforward neural
networks consisting of an input layer, hidden layer, and output layer,
as shown in Figure \ref{fig:algorithm}b.
Weight parameters between the input and hidden
layers are denoted as $\bm{\alpha}$, and those between the hidden and
output layers are $\bm{\beta}$.
Prediction is done by $G(\bm{x_i} \bm{\alpha} + \bm{b}) \bm{\beta}$,
where $G$ is an activation function, $\bm{x_i}$ is an $n$-dimensional
input data at time $i$, and $\bm{b}$ is a bias vector.
As for the training phase, $\bm{\alpha}$ is initialized with random
numbers once, while $\bm{\beta}$ is sequentially updated for every
incoming data using preceding values.
That is, at time $i$, $\bm{\beta}$ is updated with preceding
parameters $\bm{\beta_{i-1}}$, temporary vector $\bm{P_i}$, and hidden
layer output $\bm{H_i}$ calculated by input data $\bm{x_i}$, as shown
in Figure \ref{fig:algorithm}c.
Such sequential processing is suited to IoT devices with limited
memory capacity since only a single data item is retained at a time.

For semi-supervised anomaly detection, OS-ELM is combined with an
autoencoder \cite{Science2006_Hinton}, where the numbers of input
nodes $n$ and output nodes $m$ are the same.
At the training phase, $\bm{\beta}$ is updated to minimize the loss
defined by $L(\bm{y}, \bm{t})$, where $L$ is a loss function, $\bm{y}$
is a neural network output, $\bm{t}$ is a normal data, and $\bm{t} = \bm{x}$.
It is then used for anomaly detection.
Since it has been trained with normal data, the reconstruction error
$L(\bm{y}, \bm{x})$ is interpreted as an anomaly score of an input
data $\bm{x}$.
Only normal data is needed at the training phase, enabling the
ODL approach to be adapted to a deployed environment.
In practice, more than $N$ normal samples should be trained in a
deployed environment so that it can be used for anomaly detection
that is specialized for the environment, where $N$ is the number of
hidden layer nodes.

Although $\bm{x}$ could be a mini-batch of input data, its
size is restricted to one as in \cite{TC2020_Tsukada}, so that a pseudo
inverse operation needed in the sequential training algorithm (blue
expression in Figure \ref{fig:algorithm}c) is replaced with a
reciprocal operation.
In this case, the prediction and sequential training are implemented
with vector addition, vector multiplication, and scalar division,
eliminating the need for a numerical library for a pseudo inverse
operation or singular value decomposition (SVD).
In this article, it is implemented with C/C++ on a Raspberry Pi Pico.

OS-ELM based models including our ODL approach 
\cite{TC2020_Tsukada} basically have a single hidden layer that can be
trained, so their expressive power is limited compared with deep neural
networks (DNNs).
Nonetheless, practical anomaly detection tasks often deal with complex
patterns containing multiple normal/anomalous modes.
For example, a cooling fan in Figure \ref{fig:noisy_env} has three
speed levels.
If we assume they are all normal, the anomaly detection system should
recognize three different patterns as normal patterns, requiring a
rich expressive power.

To address this issue, in this article multiple ODL
instances are created for multiple normal patterns to form a multiple
classification system (MCS) \cite{FUSION2014_Wozniak}, in which a
single instance is specialized for a single pattern and a set of them
altogether handles multiple patterns.
If we have $K$ instances, 
\begin{enumerate}
\item Prediction is done by $K$ instances to produce loss values using their own parameters. The smallest loss value among them is used as an anomaly score, as shown in Figure \ref{fig:algorithm}b.
\item Sequential training is performed by a single instance that outputs the smallest loss value, as shown in Figure \ref{fig:algorithm}c.
\end{enumerate}
An instance with the smallest loss value is interpreted as the closest
instance to a given pattern.
By repeating this sequential training process, the instances are
specialized to a number of specific patterns.
Note that an initial clustering is done by a sequential k-means algorithm
during an initial training phase.

ODL has two modes: predict and train (Figure \ref{fig:algorithm}a).
In the predict mode, weight parameters are not updated.
These modes are switched either manually or automatically.
The manual mode change is triggered by a train button; users or
field-engineers push the train button when retraining is needed,
indicating the current input data is the new normal.
The initial training and clustering rely on the manual training.
The modes can also be switched automatically when pre-specified concept drifts,
such as sudden, gradual, incremental, and reoccurring ones \cite{TKDE2019_Lu},
are detected by algorithms.
For example, a fully-sequential concept drift detection algorithm 
\cite{PAISE2023_Yamada} can also be used in conjunction with the ODL algorithm
toward the autonomous retraining.

\section{IMPLEMENTATION}

We built an edge-cloud anomaly detection system as shown in Figure
\ref{fig:edge-cloud}.
Figure \ref{fig:prototype} illustrates implementations of our edge
devices, where four ODL instances are implemented on a
Raspberry Pi Pico (consisting of an ARM Cortex-M0+ CPU at 133 MHz and
264 KiB SRAM).
We use the GCC cross compiler for ARM Cortex-R/M processors v6.3.1, and
the optimization level is -O3.
The edge devices are equipped with an accelerometer or a thermal
camera, and those with the accelerometer are used in this
article.
They are attached to target objects using a magnet to observe
vibration patterns on the targets.
Anomaly detection results on the vibration patterns are sent to a
wireless gateway with LoRa as an LPWA technology.
They are then transferred to the cloud so that users can monitor the
results via a web interface as shown in Figure \ref{fig:edge-cloud}.
Grafana and MySQL are used for the visualization and database,
respectively, at the cloud side.
In the edge devices, STMicroelectronics STM32WLE5JC SoC (consisting of
an ARM Cortex-M4 CPU at 48 MHz) that supports LoRa modulation is used for
the wireless communication.
These edge devices are battery-powered in our system, so their power
consumption determines their lifetime.

\begin{figure*}[t]
\centerline{\includegraphics[width=1.1\linewidth]{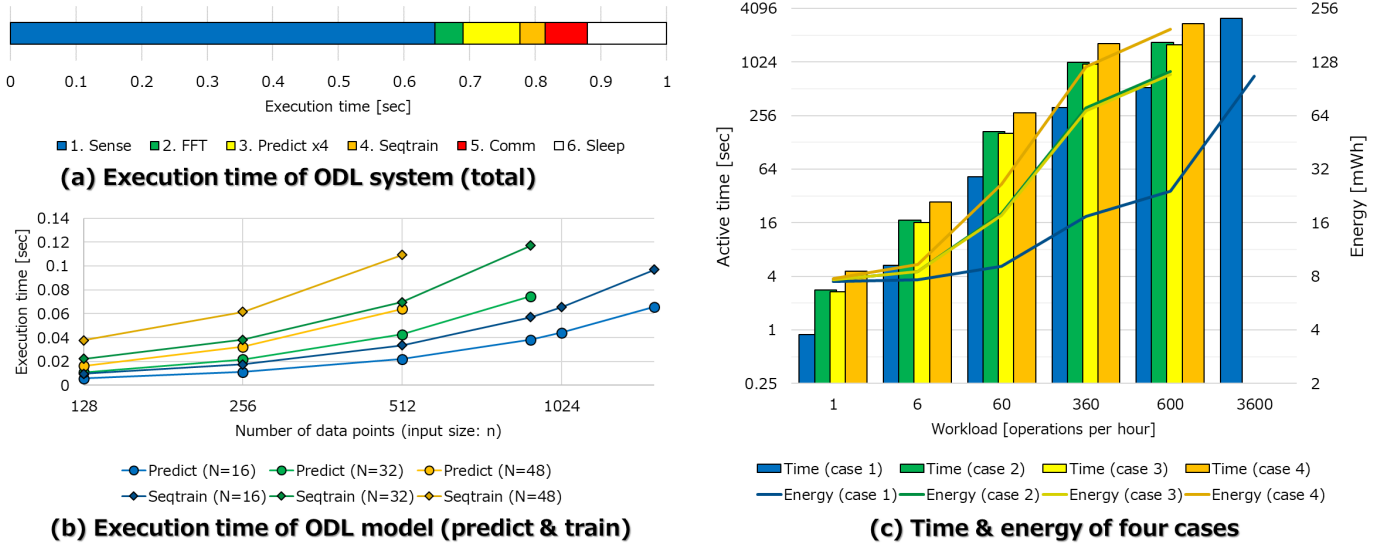}}
\caption{Execution time and energy consumption.
  ODL is analyzed in terms of execution time (left), and it is
  compared with other approaches in terms of time and energy (right).}
\label{fig:time_and_energy}
\end{figure*}

Figure \ref{fig:time_and_energy}a shows the execution time breakdown
at the edge side when anomaly detection is executed every second.
The breakdown consists of the following six parts:
\begin{enumerate}
\item A 1024-point acceleration data is received from an accelerometer
  via Serial Peripheral Interface (SPI) at 2 MHz.
  It is referred to as sensing.
\item Fast Fourier transformation (FFT) and downsampling are executed
  to produce a 256-point frequency spectrum ranging from 1 to 512 Hz
  at a 2 Hz resolution.
  They are referred to as preprocessing.
\item Prediction with four instances is performed to calculate their
  loss values.
\item Sequential training is performed by a single instance that
  outputs the smallest loss value.
\item A 20 B anomaly detection result containing an anomaly score $l$,
  output class $k$, and headers is transmitted to a gateway via
  LoRa (see Figure \ref{fig:edge-cloud}).
  It is referred to as communication.
\item Deep sleep mode of Raspberry Pi Pico.
\end{enumerate}

The four ODL instances each with $n = 256$, $N = 32$, and $m = 256$
are implemented on the Raspberry Pi Pico.
Sigmoid and mean squared error are used as an activation function $G$
and loss function $L$, respectively.
$\bm{\alpha}$ is shared by all four instances, while $\bm{P}$ and
$\bm{\beta}$ are individual for each instance.
The parameter size is thus $nN + 4NN + 4Nm$ in total.
Assuming float32 is used as a number format, the memory usage is 176
KiB, which can be implemented in the 264 KiB SRAM of the Raspberry Pi
Pico.
Execution times of the prediction and sequential training of possible
configurations are shown in Figure \ref{fig:time_and_energy}b.
The execution time of a single prediction is shorter than that of a
sequential training while it is executed four times when the number of
ODL instances is four.
A larger $N$ enriches the expressive power while it consumes more
memory and thus limiting the input size $n$.
In any cases, their execution times do not overwhelm the others, as
shown in Figure \ref{fig:time_and_energy}a.

Power consumption of the edge device is measured by a TI INA219 current
sensor at 600 samples per second.
The Raspberry Pi Pico consumes 104.5 to 117.3 mW depending on a given
workload, while it is only 6.9 mW in the deep sleep mode.
For the LPWA communication, we use the 923 MHz LoRa channel as a
license-free radio band in our region.
The spreading factor is 7, channel bandwidth is 125 kHz, and
transmission power is 20 mW to cover hundreds meter distance
with an approximately 5.47 kbps bitrate.
In this configuration, the LoRa module consumes 174.1 mW when
transmitting.
The power consumption becomes very small in the sleep mode.
In our following experiment, we assume the state is properly changed
to the sleep mode when it is not used, which is a preferred behavior
in wireless sensor nodes.

\section{EXPERIMENTAL RESULTS}

\subsection{Energy and Execution Time}

The following four cases are examined to see energy and execution time
benefits of ODL.
\begin{itemize}
\item Case 1: ODL that performs sensing, FFT,
  prediction, and sequential training at edge.
  Prediction result (20 B) is transmitted to a gateway with LoRa.
\item Case 2: Prediction-only edge AI that performs sensing, FFT, and
  prediction at edge.
  In addition to the prediction result, preprocessed data after FFT
  (1024 B) is also transmitted to the gateway if needed for a
  retraining purpose at the cloud.
\item Case 3: Vibration sensor that performs sensing and FFT at edge.
  Preprocessed data after FFT (1024 B) is transmitted to the gateway.
\item Case 4: Acceleration sensor that performs only sensing at edge.
  Raw acceleration data (2048 B) is transmitted to the gateway.
\end{itemize}
ODL (case 1) can reduce the communication size compared with the other
cases.

\begin{figure*}[t]
\centerline{\includegraphics[width=1.0\linewidth]{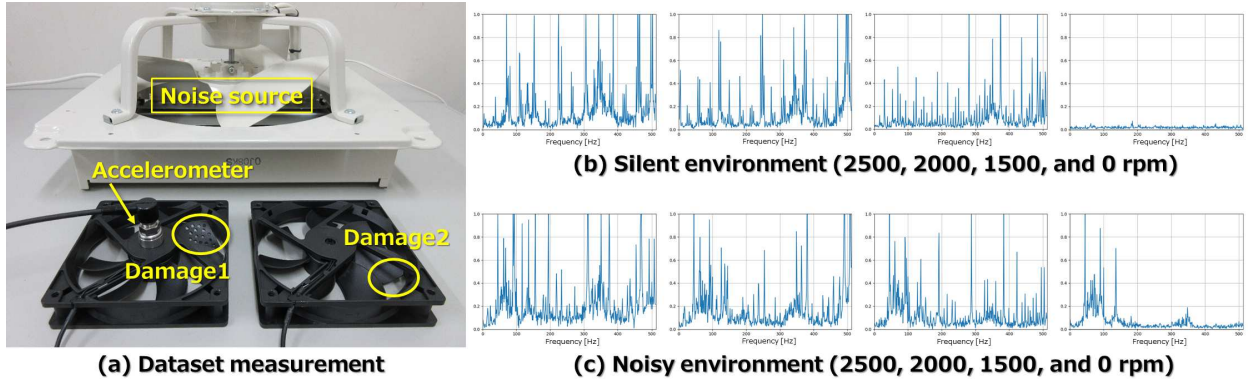}}
\caption{Cooling fan dataset containing vibration patterns of
  different speeds in silent and noisy environments (right).
  They were measured with and without noises (left).}
\label{fig:noisy_env}
\end{figure*}

Figure \ref{fig:time_and_energy}c shows the evaluation results of the
four cases.
The X-axis indicates the number of operations per hour as workload.
The left and right Y-axes are the active execution time (not including
the sleep time) in seconds and energy consumption in milliwatts per
hour (mWh), respectively.
They are log-scaled.
We assume that case 2 transfers preprocessed data for future
retraining at the cloud to align with case 1, which can retrain.
The necessity of the retraining in a deployed environment will be
demonstrated later in this section.
In all cases, their time and energy differences are not significant at
low workloads (e.g., once per hour).
As the workload is increased to once per minute or more, the time and
energy for wireless communication become dominant in cases other than
ODL.
Only ODL (blue bars and line) can implement the per-second operation
while providing a retraining capability to adapt to a new environment,
which will be evaluated next.

\subsection{Accuracy}

We evaluate the benefits of ODL when target environments around the
edge devices are changed.
The ODL model that can be retrained in a given environment is
compared with prediction-only models of DNN and OS-ELM in terms of
prediction accuracy.
The DNN models here have three hidden layers and are trained by
a backpropagation algorithm.
They are implemented in C/C++.

\subsubsection{Experimental Setup}

\begin{table*}[t]
\centering
\caption{DNN and ODL models used in accuracy evaluation.
  DNN models consume more memory for mini-batch training compared with
  on-device sequential training.
  Hyperparameters of DNN models are selected to maximize
  prediction accuracy.}
\label{tb:acc_models}
\begin{footnotesize}
\begin{tabular}{l|c|c|c}\hline \hline
& DNN (anomaly detection) & DNN (classification) & ODL \\ \hline
Train method & Backprop \& SGD (mini-batch) & Backprop \& SGD (mini-batch) & OS-ELM (sequential) \\
Layers \& nodes & \{512, 128, 64, 128, 512\} & \{512, 256, 96, 16, 4\} & \{512, 64, 512\} $\times$ inst\_num \\
Hyperparameters & batch: 5, epoch: 5, learn\_rate: 0.002 & batch: 24, epoch: 5, learn\_rate: 0.001 & inst\_num: 4 \\
Input data memory & 600 KiB & 2400 KiB & 2 KiB \\
Weights \& features & 1748 KiB & 1854 KiB & 733 KiB \\ \hline
\end{tabular}
\end{footnotesize}
\end{table*}

\begin{figure*}[t]
\centerline{\includegraphics[width=1.1\linewidth]{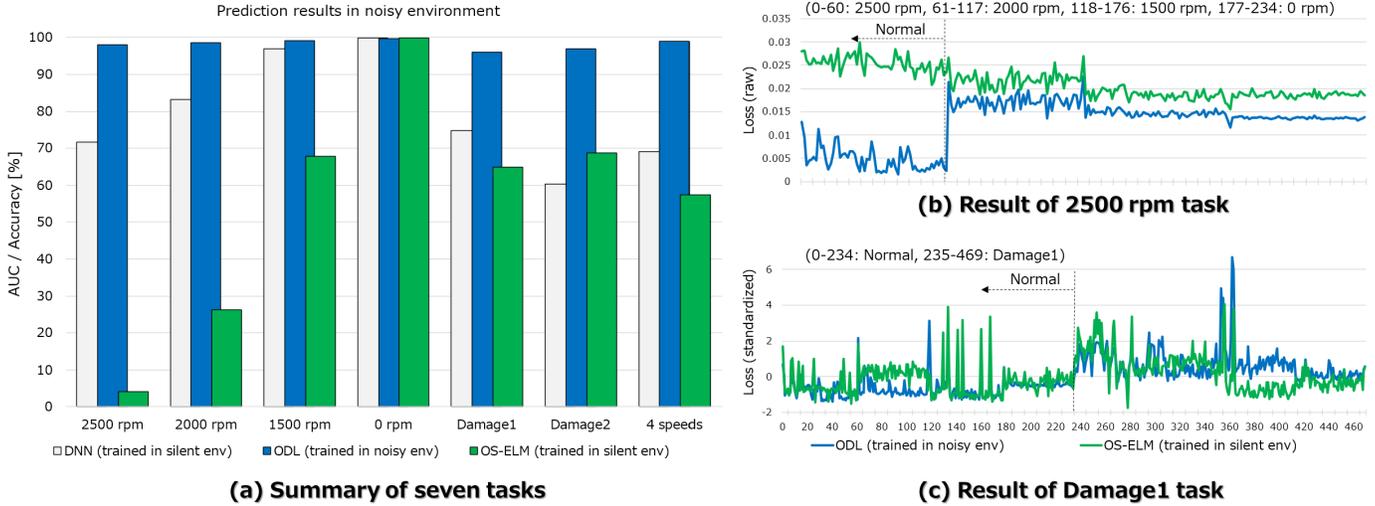}}
\caption{Accuracy of ODL and prediction-only
  models (DNN and OS-ELM) in noisy environment (left).
  The ODL model recognizes normal patterns accurately while
  prediction-only model does not (right).}
\label{fig:accuracy}
\end{figure*}

In this experiment, we use the vibration patterns of a cooling fan
dataset captured by a PCB M607A11 accelerometer, as shown in Figure
\ref{fig:noisy_env}a.
Normal and damaged 12 cm fans supporting three speed levels (i.e.,
2500, 2000, and 1500 rpm) are running in an office room (referred to
as a silent environment) and near a ventilation fan as a noise source
(referred to as a noisy environment), as shown in Figure \ref{fig:noisy_env}a.
Examples of their vibration patterns after FFT are shown in Figures
\ref{fig:noisy_env}b and \ref{fig:noisy_env}c.
Each contains a frequency spectrum ranging from 1 to 512 Hz.
The leftmost waveforms are those at 2500 rpm, and the rightmost ones
are at 0 rpm.
Anomaly detection is conducted in the noisy environment.
We assume the prediction-only models were trained in the silent
environment, while the ODL model can be retrained in the
deployed environment.
To align with the cooling fan dataset, the input size of the ODL
and prediction-only models is set to 512.
Their model parameters are listed in Table \ref{tb:acc_models}.
The hyperparameters of the DNN models, such as batch size and number
of epochs, were selected so as to maximize the prediction accuracy.

The cooling fan dataset containing more than 11,000 waveforms is
publicly available on
GitHub \footnote{\url{https://github.com/matutani/cooling-fan}}.
The following seven tasks are used for the accuracy evaluation.
\begin{itemize}
\item 2500 rpm: Fan speeds are changed among 2500, 2000, 1500,
  and 0 rpm in the noisy environment.
  The task is to detect different fan speeds other than 2500 rpm as
  anomalous.
\item 2000 rpm: Same conditions as those for 2500 rpm, but the task
  is to detect fan speeds other than 2000 rpm as anomalous.
\item 1500 rpm: Same conditions as those for 2500 rpm, but the task
  is to detect fan speeds other than 1500 rpm as anomalous.
\item 0 rpm: Same conditions as those for 2500 rpm, but the task
  is to detect fan speeds other than 0 rpm as anomalous.
\item Damage1: Normal fan operates first, and then an unbalanced
  fan with holes (Damage1 in Figure \ref{fig:noisy_env}a) operates
  instead.
  The task is to detect the damaged one as anomalous.
\item Damage2: Same as Damage1 but an unbalanced fan with a chipped
  blade (Damage2 in Figure \ref{fig:noisy_env}a) is used to be
  detected as anomalous.
\item 4 speeds: Same conditions as those for 2500 rpm, but the task is
  to classify four speeds to see classification accuracy.
\end{itemize}
For the 2500 rpm task, 300 samples and 235 samples were used for
training and prediction, respectively, for each test scenario.
Multiple tests were executed both for the ODL
and the prediction-only cases using different samples.
The same test procedure was applied to the 2000, 1500, and 0 rpm
tasks.
For the Damage1 task, 1200 samples and 470 samples were used for
training and prediction, respectively for each test scenario.
The same procedure was applied to the Damage2 and 4 speeds tasks.

The 4 speeds task is evaluated with classification accuracy while
the other tasks are with an area under the receiver operating
characteristic curve (AUC), which is calculated with anomaly scores $l$.
The classification accuracy is calculated with classes $k$ predicted
by the ODL instances.

\subsubsection{Results and Discussion}

Figure \ref{fig:accuracy}a shows the evaluation results of the ODL
(blue bars) and prediction-only models (DNN in white bars and OS-ELM
in green bars) with the seven tasks.
The Y-axis is the classification accuracy for the 4 speeds task, and
the AUC for the other tasks.
Detecting 0 rpm as normal is an easy task for all the approaches even
in the noisy environment, while in the 2500 rpm task, the
prediction-only models fail to detect 2500 rpm as normal in the noisy
environment.
Figure \ref{fig:accuracy}b illustrates a part of the result in the
2500 rpm task, in which samples 0-60 are normal (i.e., 2500 rpm) and the
others are anomalous (i.e., 2000, 1500, or 0 rpm).
The ODL model (blue line) recognizes this difference
accurately while the prediction-only model (OS-ELM in green line) does
not in the noisy environment.
For the Damage1, Damage2, and 4 speeds tasks, the ODL model is also
better than the prediction-only models in the noisy environment.
Figure \ref{fig:accuracy}c illustrates a part of the result in the Damage1
task, in which samples 0-234 are normal and the others are anomalous
(i.e., Damage1) \footnote{Y-axis is standardized loss value in each
instance so that the scales of ODL and prediction-only
model outputs are adjusted to clearly show their difference.}.
As shown, the loss value of the ODL model is stably high after
the damaged fan is used, while mispredictions are observed in the
prediction-only model.
These results demonstrate the advantages of the ODL model when
there is a gap between the trained and deployed environments.

\section{SUMMARY}

In real-world anomaly detection, normal and anomalous patterns may
vary depending on a given environment.
This article introduces a neural network based ODL approach to address
this issue.
Our approach is the semi-supervised sequential training of
multiple neural networks tailored for low-end IoT devices, such as
wireless sensor nodes.
The experimental results demonstrate that retraining by the ODL
model improves the accuracy compared with a prediction-only DNN model when
there is a gap between the trained and deployed environments.
It also saves the communication cost and energy consumption for
battery-powered IoT devices.
A demonstration video of the proposed ODL approach on 
wireless sensor nodes is available on
YouTube \footnote{\url{https://youtu.be/xCQNZ7AuB-M}}.
Our future plan includes an ODL chip design and its
application development.

{\bf Acknowledgements}
This work was supported in part by JST AIP Acceleration Research 
JPMJCR23U3, Japan. 
The authors would like to thank Dr. Mineto Tsukada for discussions.


\begin{thebibliography}{1}

\bibitem{IEEE2019_Zhou}
Zhi Zhou et~al.
\newblock {Edge Intelligence: Paving the Last Mile of Artificial Intelligence
  With Edge Computing}.
\newblock {\em Proc. of IEEE}, 107(8):1738--1762, 2019.

\bibitem{TKDE2019_Lu}
Jie Lu et~al.
\newblock {Learning under Concept Drift: A Review}.
\newblock {\em IEEE Trans. on Knowledge and Data Engineering},
  31(12):2346--2363, 2019.

\bibitem{TC2020_Tsukada}
Mineto Tsukada et~al.
\newblock {A Neural Network-Based On-device Learning Anomaly Detector for Edge
  Devices}.
\newblock {\em IEEE Trans. on Computers}, 69(7):1027--1044, 2020.

\bibitem{TNN2006_Liang}
{Nan-ying} Liang et~al.
\newblock {A Fast and Accurate Online Sequential Learning Algorithm for
  Feedforward Networks}.
\newblock {\em IEEE Trans. on Neural Networks}, 17(6):1411--1423, 2006.

\bibitem{Science2006_Hinton}
G.~Hinton et~al.
\newblock {Reducing the Dimensionality of Data with Neural Networks}.
\newblock {\em Science}, 313(5786):504--507, 2006.

\bibitem{FUSION2014_Wozniak}
Michal Wozniak et~al.
\newblock {A Survey of Multiple Classifier Systems as Hybrid Systems}.
\newblock {\em Information Fusion}, 16:3--17, 2014.

\bibitem{PAISE2023_Yamada}
Takeya Yamada et~al.
\newblock {A Lightweight Concept Drift Detection Method for On-Device Learning
  on Resource-Limited Edge Devices}.
\newblock In {\em {Proc. of IPDPS Workshops}}, pages 761--768, May 2023.

\end{thebibliography}

\end{document}